\documentclass[10pt,twocolumn,letterpaper]{article}

\usepackage{cvpr}
\usepackage{times}
\usepackage{epsfig}
\usepackage{graphicx}
\usepackage{amsmath}
\usepackage{amssymb}
\usepackage{amsthm}
\usepackage{float}
\usepackage{capt-of}

\usepackage[pagebackref=true,breaklinks=true,letterpaper=true,colorlinks,bookmarks=false]{hyperref}

\usepackage[final]{pdfpages}
\cvprfinalcopy %

\def\cvprPaperID{} %

\input{macros}

\usepackage{enumitem}
\setitemize{noitemsep,topsep=0pt,parsep=0pt,partopsep=0pt}

\ifcvprfinal\fi
\begin{document}

\title{FML: Face Model Learning from Videos}

\author{Ayush Tewari$^1$~~~Florian Bernard$^1$~~~Pablo Garrido$^2$~~~Gaurav Bharaj$^2$~~~Mohamed Elgharib$^1$ \\ \vspace{0.4cm}
Hans-Peter Seidel$^1$~~~Patrick P{\'e}rez$^3$~~~Michael Zollh{\"o}fer$^4$~~~Christian Theobalt$^1$\\ 
		$^1$MPI Informatics, Saarland Informatics Campus~~~$^2$Technicolor~~~$^3$Valeo.ai~~~$^4$Stanford University
}

\twocolumn[{
	\renewcommand\twocolumn[1][]{#1}
	\maketitle
	\begin{center}
		\vspace{-0.8cm}
\includegraphics[width=\linewidth]{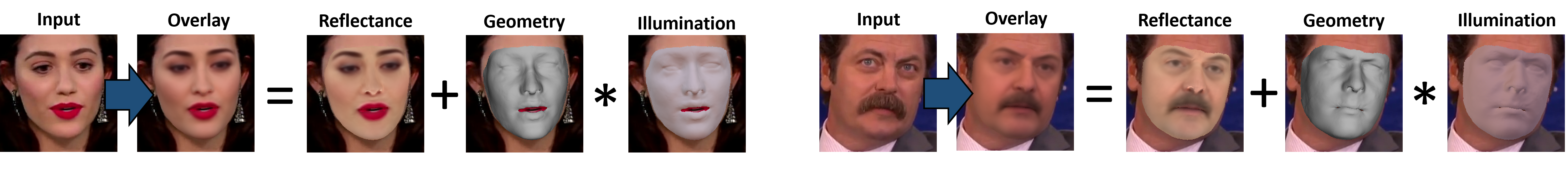}
		\vspace{-0.5cm}
		\captionof{figure}
		{
			We propose multi-frame self-supervised training of a deep network based on in-the-wild video data for jointly learning a face model and 3D face reconstruction.
			Our approach successfully disentangles facial shape, appearance, expression, and scene illumination.
		}
		\label{fig:teaser}
	\end{center}

	\vspace{0.1cm}
}]

\begin{abstract}
Monocular image-based 3D reconstruction of faces is a long-standing problem in computer vision. Since image data is a 2D projection of a 3D face, the resulting depth ambiguity makes the problem ill-posed. Most existing methods rely on data-driven priors that are built from limited 3D face scans. In contrast, we propose multi-frame video-based self-supervised training of a deep network that (i) learns a face identity model both in shape and appearance while (ii) jointly learning to reconstruct 3D faces. Our face model is learned using only corpora of \it{in-the-wild} video clips collected from the Internet. This virtually endless source of training data enables learning of a highly general 3D face model. In order to achieve this, we propose a novel multi-frame consistency loss that ensures consistent shape and appearance across multiple frames of a subject's face, thus minimizing depth ambiguity. At test time we can use an arbitrary number of frames, so that we can perform both monocular as well as multi-frame reconstruction. 
\end{abstract}

\section{Introduction}
The reconstruction of faces from visual data 
has a wide range of applications in vision and graphics, including face tracking, emotion recognition, and interactive image/video editing tasks relevant in multimedia. 
Facial images and videos are ubiquitous, as smart devices as well as consumer and professional cameras provide a continuous and virtually endless source thereof. %
When such data is captured without controlled scene location, lighting, or intrusive equipment (e.g. egocentric cameras or markers on actors), one speaks of {\it ``in-the-wild''} images.
Usually {\it in-the-wild} data is of low resolution, noisy, or contains motion and focal blur, making the reconstruction problem much harder than in a controlled setup.
3D face reconstruction from {\it in-the-wild} monocular 2D image and video data \cite{Zollhoefer2018FaceSTAR} deals with disentangling facial shape identity (neutral geometry), skin appearance (or albedo) and expression, %
as well as estimating the scene lighting and camera parameters.
Some of these attributes, \eg albedo and lighting, are not easily separable in monocular images. Besides, poor scene lighting, depth ambiguity, and occlusions due to facial hair, sunglasses and large head rotations complicates 3D face reconstruction.

In order to tackle the difficult monocular 3D face reconstruction problem, most existing methods rely on the availability of strong prior models that serve as regularizers for an otherwise ill-posed problem \cite{Blanz2003,ekman1997face,Vlasic:2005:FTM}.
Although such approaches achieve impressive facial shape and albedo reconstruction, %
they introduce an inherent bias due to the used face model. For instance, the 3D Morphable Model (3DMM) by Blanz et al.~\cite{Blanz2003} is based on a comparably small set of 3D laser scans of Caucasian actors, thus limiting generalization to general real-world identities and ethnicities.
With the rise of CNN-based deep learning, various techniques have been proposed, which in addition to 3D reconstruction also perform face model learning from monocular images \cite{Tran18,Tran2018b,tewari2018self,NeuralFace2017}. %
However, these methods heavily rely on a pre-existing 3DMM to resolve the inherent depth ambiguities of the monocular reconstruction setting.
Another line of work, where 3DMM-like face models are not required, are based on photo-collections \cite{Kemelmacher2013,Liang16,Suwajanakorn2014}.
However, these methods need a very large number (\eg ${\approx}100$) of facial images of the same subject, and thus they impose strong demands on the training corpus.

In this paper, we introduce an approach that learns a comprehensive face identity model using clips crawled from \emph{in-the-wild} Internet videos \cite{Chung18b}. This face identity model comprises two components: One component to represent the geometry of the facial identity (modulo expressions), and another to represent the facial appearance in terms of the albedo.
As we have only weak requirements on the training data (cf.~Sec.~\ref{sec:data}), our approach can employ a virtually endless amount of community data and thus obtain a model with better generalization; laser scanning a similarly large group of people for model building would be nearly impossible. 
Unlike most previous approaches, we do not require a pre-existing shape identity and albedo model as initialization, but instead learn their variations from scratch. As such, our methodology is applicable in scenarios when no existing model is available, or if it is difficult to create such a model from 3D scans (\eg for faces of babies). 

From a technical point of view, one of our main contributions is a novel multi-frame consistency loss, which ensures that the face identity and albedo reconstruction is consistent across frames of the same subject.
This way we can avoid depth ambiguities present in many monocular approaches and obtain a more accurate and robust model of facial geometry and albedo.
Moreover, by imposing orthogonality between our learned face identity model and an existing blendshape expression model, our approach automatically disentangles facial expressions from identity based geometry variations, without resorting to a large set of hand-crafted priors. 
In summary, our approach is based on the following technical contributions:
\begin{enumerate}[noitemsep,topsep=0pt,parsep=0pt,partopsep=0pt,]
\item A deep neural network that learns a facial shape and appearance space from a big dataset of unconstrained images that contains multiple images of each subject, \eg multi-view sequences, or even monocular videos.
\item Explicit blendshape and identity separation by a projection onto the blendshapes' nullspace that enables a multi-frame consistency loss.
\item A novel multi-frame identity consistency loss based on a Siamese network \cite{vinyals2016matching}, with the ability to handle monocular and multi-frame reconstruction.
\end{enumerate}

\section{Related Work}

The literature on 3D model learning is quite vast and we mainly review methods for reconstructing 3D face models from scanner data, monocular video data, photo-collections and a single 2D image.
An overview of the state-of-the-art in model-based face reconstruction is given in \cite{Zollhoefer2018FaceSTAR}.

\paragraph{Morphable Models from High-quality Scans:}
3DMMs represent deformations in a low-dimensional subspace and are often built from scanner data \cite{Blanz1999,Bogo14,Li17}.
Traditional 3DMMs model geometry/appearance variation from limited data via PCA \cite{Blanz1999,Blanz2003,Hasler09}.
Recently, richer PCA models have been obtained from large-scale datasets \cite{booth20163d,Pishchulin17}.
Multilinear models generalize statistical models by capturing a set of mutually orthogonal variation modes (e.g., global and local deformations) via a tensor decomposition \cite{Vlasic:2005:FTM,Bolkart15,Bolkart16}.
However, unstructured subspaces or even tensor generalizations are incapable of modeling localized deformations from limited data.
In this respect, Neumann et al.~\cite{Neumann13} and Bernard et al.~\cite{bernard2016linear} devise methods for computing sparse localized deformation components directly from mesh data.
L\"{u}thi et al. \cite{Luethi18} propose the so-called Gaussian Process morphable models (GPMMs), which are modeled with arbitrary non-linear kernels, to handle strong non-linear shape deformations.
Ranjan et al. \cite{Ranjan18} learn a non-linear model using a deep mesh autoencoder with fast spectral convolution kernels.
Garrido et al. \cite{GarridoZWBPBT16} train radial basis functions networks to learn a corrective 3D lip model from multiview data.
In an orthogonal direction, Li et al. \cite{Li17} learn a hybrid model that combines a linear shape space with articulated motions and semantic blendshapes.
All these methods mainly model shape deformations and are limited to the availability of scanner data.

\paragraph{Parametric Models from Monocular Data:}
Here, we distinguish between personalized, corrective, and morphable model learning.
Personalized face models have been extracted from monocular video by first refining a parametric model in a coarse-to-fine manner (e.g., as in \cite{Roth:2016}) and then learning a mapping from coarse semantic deformations to finer non-semantic detail layers \cite{Ichim15,GarriZCVVPT2016}.
Corrective models represent out-of-space deformations (e.g., in shape or appearance) which are not modeled by the underlying parametric model.
Examples are adaptive linear models 
customized over a video sequence \cite{Bouaziz13,Hsieh15} or non-linear models learned from a training corpus \cite{Richardson_2017_CVPR,tewari2018self}.
A number of works have been proposed for in-the-wild 3DMM learning
\cite{Sengupta18,Tran18,Bas18,Booth_2017_CVPR}.
Such solutions decompose the face into its intrinsic components
through encoder-decoder architectures that exploit weak supervision.
Tran et al. \cite{Tran18} employ two separate convolutional decoders to learn a non-linear model that disentangles shape from appearance.
Similarly, Sengupta et al. \cite{Sengupta18} propose 
residual blocks to produce a complete separation of surface normal and albedo features.
There also exist approaches that learn 3DMMs of rigid \cite{TulsianiKCM17} or articulated objects \cite{KanazawaTEM18} by leveraging image collections.
These methods predict an instance of a 3DMM directly from an image \cite{KanazawaTEM18} or use additional cues (e.g., segmentation and shading) to fit and refine a 3DMM \cite{TulsianiKCM17}.

\paragraph{Monocular 3D Reconstruction:}
Optimization-based reconstruction algorithms rely on a personalized model \cite{Cao16,Fyffe14,GVWT13,Wu16b} or a parametric prior \cite{Agudo14,Bouaziz13,Li13,GarriZCVVPT2016,Shi14} to estimate 3D geometry from a 2D video.
Learning-based approaches regress 3D face geometry from a single image by learning an image-to-parameter or image-to-geometry mapping \cite{olszewski2016high,Richardson_2017_CVPR,tewari17MoFA,tewari2018self,sela2017unrestricted,Tran_2017_CVPR,KimZTTRT17}.
These methods require ground truth face geometry \cite{Tran_2017_CVPR, Laine:2017}, a morphable model from which synthetic training images are generated \cite{RichaSK2016,Richardson_2017_CVPR,sela2017unrestricted,KimZTTRT17}, or a mixture of both \cite{McDonagh2016,Klaudiny:2017}.
Recently, Tewari et al.~\cite{tewari17MoFA} trained fully unsupervised through an inverse rendering-based loss. However, color and shape variations lie in the subspace of a parametric face prior.
Only very recent methods for monocular face reconstruction \cite{tewari2018self,Tran18,Tran2018b,Booth_2017_CVPR} allow for out-of-space model generalization while training from in-the-wild data.

\paragraph{3D Reconstruction via Photo-collections:}
Face reconstruction is also possible by fitting a template model to 
photo-collections.
In \cite{Kemelmacher-Shlizerman:2011}, an average shape and appearance model is reconstructed from a person-specific photo-collection via low-rank matrix factorization.
Suwajanakorn et al.~\cite{Suwajanakorn2014} use this model to track detailed facial motion from unconstrained video.
Kemelmacher-Shlizerman \cite{Kemelmacher2013} learns a 3DMM from a large photo-collection of people, grouped into a fixed set of semantic labels.
Also, Liang et al. \cite{Liang16} leverage multi-view person-specific photo-collections to reconstruct the full head.
In a different line of research, Thies et al. \cite{Thies15} fit a coarse parametric model to user-selected views to recover personalized face shape and albedo.
Roth et al.~\cite{Roth:2016} personalize an existing morphable model to an image collection by using a coarse-to-fine photometric stereo formulation.
Note that most of these methods do not learn a general face model, e.g. a shape basis that spans the range of facial shapes of an entire population, but instead they obtain a single person-specific 3D face instance.
Besides, these methods require curated photo-collections. We, on the contrary, build a 3DMM representation that generalizes across multiple face identities and impose weaker assumptions on the training data.

\paragraph{Multi-frame 3D Reconstruction:}
Multi-frame reconstruction techniques exploit either temporal information or multiple views to better estimate 3D geometry.
Shi et al. \cite{Shi14} globally fit a multilinear model to 3D landmarks at multiple keyframes and enforce temporal consistency of in-between frames via interpolation.
In \cite{GarriZCVVPT2016}, person-specific facial shape is obtained by averaging per-frame estimates of a parametric face model.
Ichim et al. \cite{Ichim15} employ a multi-view bundle adjustment approach to reconstruct facial shape and refine expressions using 
actor-specific sequences.
Piotraschke et al. \cite{PiotraschkeB16} combine region-wise reconstructions of a 3DMM from many images using a normal distance function.
Garg et al. \cite{GargRA13} propose a model-free approach that globally optimizes for dense 3D geometry in a non-rigid structure from motion framework.
Beyond faces, Tulsian et al. \cite{TulsianiZEM17} train a CNN to predict single-view 3D shape (represented as voxels) using multi-view ray consistency.

\begin{figure*}[h!t!]
	\centering
	\includegraphics[width=0.9\linewidth]{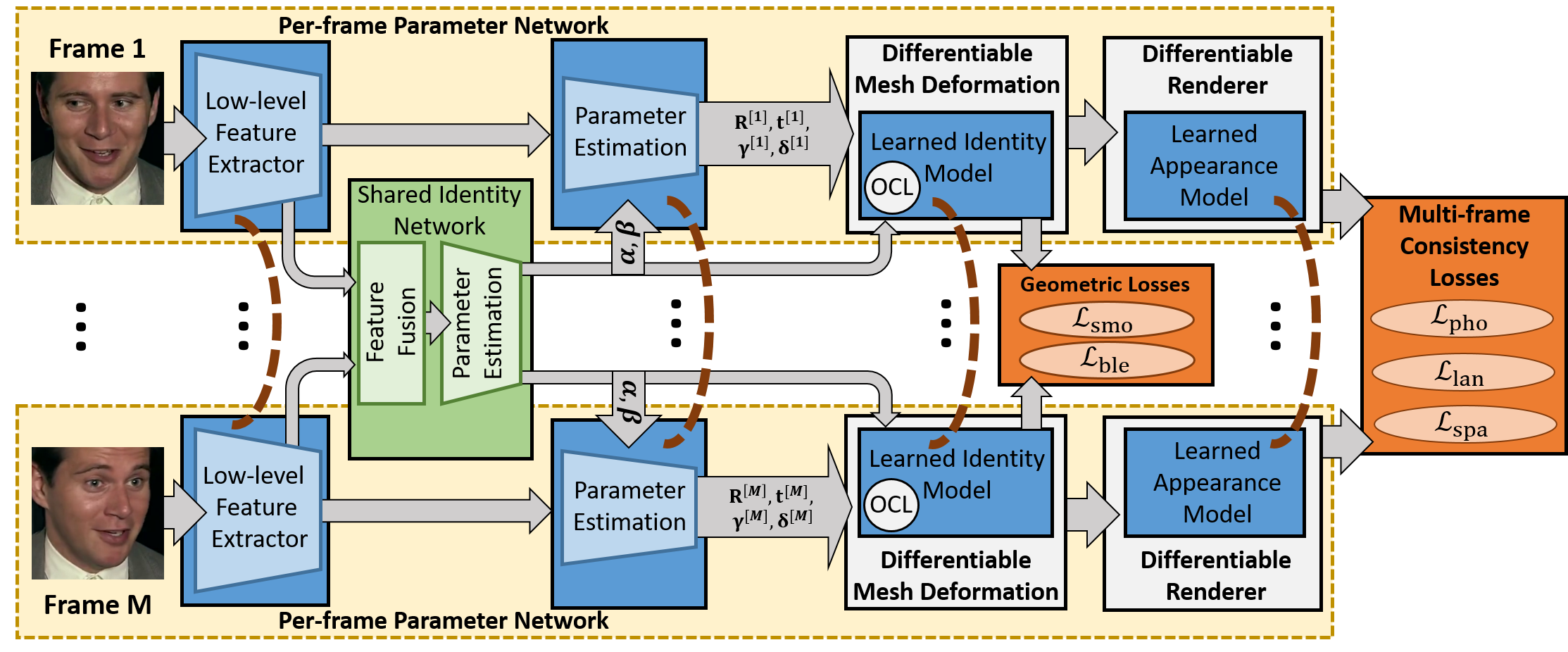}
	\caption{Pipeline overview. Given multi-frame input that shows a person under different facial expression, head pose, and illumination, our approach first estimates these parameters per frame. In addition, it jointly obtains the shared identity parameters that control facial shape and appearance, while at the same time learning a graph-based geometry and a per-vertex appearance model. We use a differentiable mesh deformation layer in combination with a differentiable face renderer to implement a model-based face autoencoder.
    }
	\label{fig:pipeline}
\end{figure*}

\section{Face Model Learning}
Our novel face model learning approach solves two tasks: it jointly learns (i) a parametric face geometry and appearance model, and (ii) an estimator for facial shape, expression, albedo, rigid pose and incident illumination parameters.
An overview of our approach is shown in Fig.~\ref{fig:pipeline}.

\paragraph{Training:}
Our network is trained in a self-supervised fashion based on a training set of multi-frame images, i.e., multiple images of the same person sampled from a video clip, see Section \ref{sec:data}.
The network jointly learns an appearance and shape identity model (Section \ref{sec:graph}). It also estimates per-frame parameters for 
the rigid head pose, illumination, and expression parameters, as well as shape and appearance identity parameters that are shared among all frames.
We train the network based on a differentiable renderer that incorporates a per-vertex appearance model and a graph-based shape deformation model (Section \ref{sec:imageformation}). To this end, we propose a set of training losses that account for geometry smoothness, photo-consistency, sparse feature alignment and appearance sparsity, see Section \ref{sec:learning}.

\paragraph{Testing:}
At test time, our network jointly reconstructs shape, expression, albedo, pose and illumination from 
an arbitrary number of face images of the same person. 
Hence, the same trained network is usable both for monocular and multi-frame face reconstruction.

\subsection{Dataset} \label{sec:data}

We train our 
approach using the VoxCeleb2 multi-frame video dataset \cite{Chung18b}.
This dataset contains over 140k videos of over 6000 celebrities crawled from Youtube.
We sample a total of $N = 404$k multi-frame images $\setF_{1},\ldots,\setF_N$ from this dataset.
The $\ell$-th multi-frame image $\setF_{\ell} = \{F^{[f]}_{\ell}\}_{f=1}^M$ comprises $M=4$ frames $F^{[1]}_{\ell},\ldots, F^{[M]}_{\ell}$ of the same person $\ell$ extracted from the same video clip to avoid unwanted variations, e.g., due to aging or accessories.
The same person can appear multiple times in the dataset.
To obtain these images, we perform several sequential steps.
First, the face region is cropped based on automatically detected facial landmarks \cite{SaragLC2011,SaragihLC11a}.
Afterwards, we discard images whose cropped region is smaller than a threshold (i.e., 200\,pixels) and that have low landmark detection confidence, as provided by the landmark tracker \cite{SaragLC2011,SaragihLC11a}.
The remaining crops are re-scaled to $240 {\times} 240$ pixels.
When sampling the $M$ frames in $\setF_{\ell}$, we ensure sufficient diversity in head pose based on the head orientation obtained by the landmark tracker.
We split our multi-frame dataset $\setF_{1},\ldots,\setF_N$ into a training (383k images) and test set (21k images).

\subsection{Graph-based Face Representation} \label{sec:graph}
We propose a multi-level face representation that is based on both a coarse shape deformation graph 
and a high-resolution surface mesh, where each vertex has a color value that encodes the facial appearance. 
This representation enables our approach to learn a face model of geometry and appearance based on multi-frame consistency.
In the following, we explain the components in detail.

\begin{figure}
	\centering
	\includegraphics[width=0.5\linewidth]{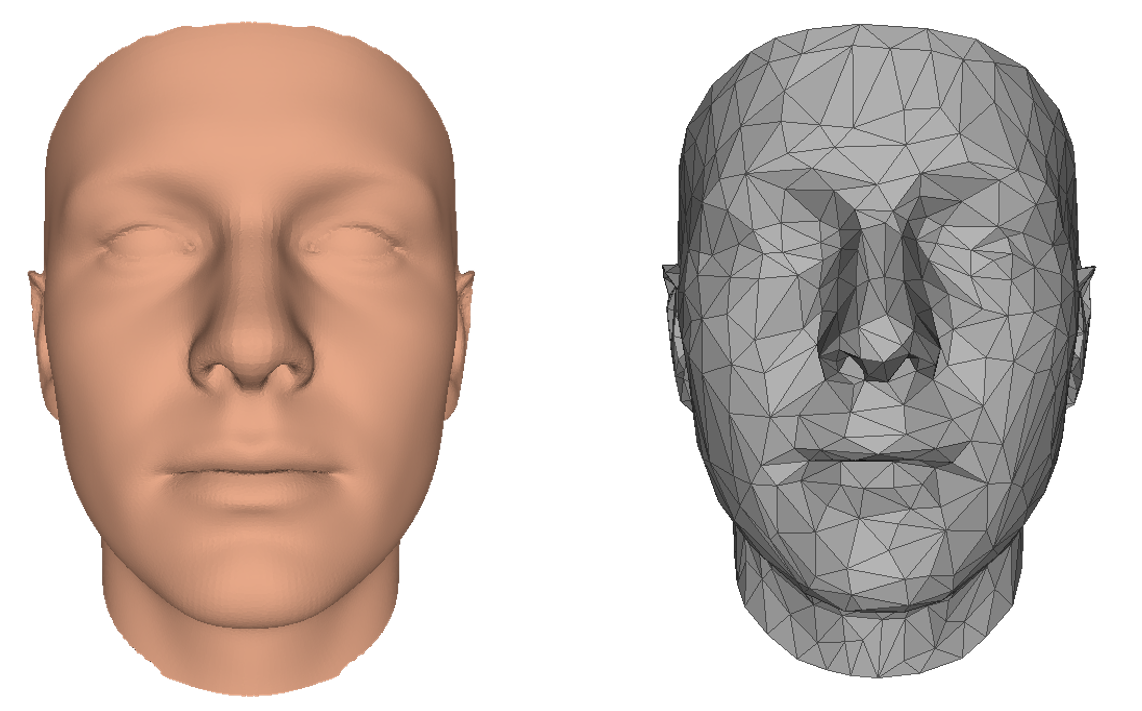}
	\caption{Neutral face shape and appearance (left), and the coarse deformation graph of the face mesh (right).}
	\label{fig:faceGraph}
	\vspace{-0.4cm}
\end{figure}

\paragraph{Learnable Graph-based Identity Model:}
Rather than learning the identity model on the high-res mesh $\setV$ with $|\setV|=60$k vertices, we simplify this task by considering a lower-dimensional parametrization based on deformation graphs~\cite{sumner2007embedded}.
We obtain our (coarse) deformation graph $\setG$ by downsampling the mesh to $|\setG|= 521$  nodes, see Fig.~\ref{fig:faceGraph}.
The network now learns a deformation on $\setG$ that is then transferred to the mesh $\setV$ via linear blend skinning.
The vector $\vecg \in \R^{3\vert \setG \vert}$ of the $|\setG|$ stacked node positions of the 3D graph is defined as %
\begin{align}
	\vecg  = \bar{\vecg} + \matTheta_s \vecalpha \,,
\end{align} 
where $\bar{\vecg} \in \R^{3\vert \setG \vert}$ denotes the mean graph node positions.
We obtain $\bar{\vecg}$ by downsampling a face mesh with slightly open mouth (to avoid connecting the upper and lower lips).
The columns of the learnable matrix $\matTheta_s \in \R^{3\vert \setG \vert \times g}$ span the $g$-dimensional ($g=500$) graph deformation subspace, and $\vecalpha \in \R^{g}$ represents the graph deformation parameters.

The vertex positions $\vecv \in \R^{3|\setV|}$ of the high-resolution mesh $\setV$ that encode the shape identity are then given as
\begin{align}\label{eq:skin}
\vecv(\matTheta_s,\vecalpha) = \bar{\vecv} + \matS \matTheta_s \vecalpha\,.
\end{align}
Here, $\bar{\vecv} \in \R^{3|\setV|}$ is fixed to the neutral mean face shape as defined in the 3DMM \cite{Blanz1999}.
The skinning matrix $\matS \in \R^{3|\setV| \times 3\vert \setG \vert}$ is obtained based on the mean shape $\bar{\vecv}$ and mean graph nodes $\bar{\vecg}$.

To sum up, our identity model is represented by a deformation graph $\setG$, where the deformation parameter $\vecalpha$ is regressed by the network while
learning the deformation subspace basis $\matTheta_s$. We regularize this ill-posed learning problem by exploiting multi-frame consistency.

\paragraph{Blendshape Expression Model:}
For capturing facial expressions, we use a linear blendshape model that combines the facial expression models from~\cite{Alexander2009} and \cite{Cao2014b}.
This model is fixed, i.e. not learned. Hence, the expression deformations are directly applied to the high-res mesh. The vertex positions of the high-res mesh that account for shape identity as well as the facial expression are given by
\begin{align}\label{eq:blend}
\vecv(\matTheta_s,\vecalpha,\vecdelta) = \bar{\vecv} + \matS \cdot\operatorname{OCL}(\matTheta_s) \vecalpha + \matB \vecdelta\,,
\end{align}
where $\matB \in \R^{3|\setV| \times b}$ is the fixed blendshape basis, %
 $\vecdelta \in \R^b$ is the vector of $b=80$ blendshape parameters,
and $\operatorname{OCL}$ is explained next.

\paragraph{Separating Shape and Expression:}
We ensure a separation of shape identity from facial expressions by imposing orthogonality between our learned shape identity basis and the fixed blendshape basis.
To this end, we first represent the blendshape basis $\matB \in \R^{3|\setV| \times b}$ with respect to the deformation graph domain %
by solving
	$\matB = \matS \matB_{\setG}$
for the graph-domain blendshape basis $\matB_{\setG} \in \R^{3|\setG| \times b_{\setG}}$ in a least-squares sense. Here, $b_{\setG} = 80$ is fixed. Then, we orthogonalize the columns of $\matB_{\setG}$.
We propose the \emph{Orthogonal Complement Layer (OCL)} to ensure that our learned $\operatorname{OCL}(\matTheta_s)$ fulfills the orthogonality constraint $\matB_{\setG} ^T\operatorname{OCL}(\matTheta_s)  = \boldsymbol{0}$. Our layer is defined in terms of the projection of $\matTheta_s$ onto the orthogonal complement $\matB_{\setG}^{\perp}$ of $\matB_{\setG}$, i.e.,
\begin{align}
\operatorname{OCL}(\matTheta_s) &= \operatorname{proj}_{\matB_{\setG}^{\perp}}(\matTheta_s) 
= \matTheta_s - \operatorname{proj}_{\matB_{\setG}}(\matTheta_s) \\
                                &= \matTheta_s - \matB_{\setG} (\matB_{\setG}^T \matB_{\setG})^{-1} \matB_{\setG}^T \matTheta_s\,.
\end{align}
The property $\matB_{\setG}^T \operatorname{OCL}(\matTheta_s) = \boldsymbol{0}$ can easily be verified.

\paragraph{Learnable Per-vertex Appearance Model:}
The facial appearance is encoded in the $3|\mathcal{V}|$-dimensional vector 
\begin{equation}
\vecr(\vecbeta) = \bar{\vecr} + \matTheta_a  \vecbeta
\end{equation}
that stacks all $|\mathcal{V}|$ per-vertex colors represented as RGB triplets.
The mean facial appearance $\bar{\vecr} \in \R^{3|\setV|}$ and the appearance basis $\matTheta_a \in \R^{3|\setV| \times |\vecbeta|}$ are learnable, while the facial appearance parameters $\vecbeta$ are regressed. Note that we initialize the mean appearance $\bar{\vecr}$ to a constant skin tone 
and 
define the reflectance directly on the high-res mesh $\setV$.

\subsection{Differentiable Image Formation} \label{sec:imageformation}
To enable end-to-end self-supervised training, we employ a differentiable image formation model that maps 3D model space coordinates $\mathbf{v} \in \mathbb{R}^3$ onto 2D screen space coordinates $\mathbf{u} \in \mathbb{R}^2$.
The mapping is implemented as
$\mathbf{u} = \Pi(\Phi(\mathbf{v})),$
where $\Phi$ and $\Pi$ denote the rigid head pose and camera projection, respectively.
We also apply a differentiable illumination model that transforms illumination parameters $\boldsymbol{\gamma}$ as well as per-vertex appearance $\mathbf{r}_i$ and normal $\mathbf{n}_i$ into shaded per-vertex color $\mathbf{c}_i(\mathbf{r}_i, \mathbf{n}_i, \boldsymbol{\gamma})$.
We explain these two models in the following.

\paragraph{Camera Model:}
We assume w.l.o.g. that
the camera space corresponds to world space.
We model the head pose via a rigid mapping $\Phi(\mathbf{v}) = \mathbf{R} \mathbf{v} + \mathbf{t}$, defined by the global rotation $\mathbf{R} \in SO(3)$ and the translation $\mathbf{t} \in \mathbb{R}^3$.
After mapping a vertex from model space $\mathbf{v}$ onto camera space $\mathbf{\hat v} = \Phi(\mathbf{v})$, the full perspective camera model $\Pi: \mathbb{R}^3 \rightarrow \mathbb{R}^2$ projects the points $\mathbf{\hat v}$ into screen space $\mathbf{u}= \Pi(\mathbf{\hat v}) \in \mathbb{R}^2$.

\paragraph{Illumination Model:}
Under the assumption of distant smooth illumination and purely \textit{Lambertian} surface properties, we employ Spherical Harmonics (SH) \cite{Ramamoorthi2001}
to represent the incident radiance at a vertex $\mathbf{v}_i$ with normal $\mathbf{n}_i$ and appearance $\mathbf{r}_i$ as
\begin{equation}
\mathbf{c}_i(\mathbf{r}_i, \mathbf{n}_i, \boldsymbol{\gamma}) = \mathbf{r}_i \cdot \sum_{b=1}^{B^2}{\boldsymbol{\gamma}_b \cdot H_b(\mathbf{n}_i)}
\enspace{.}
\end{equation}
The illumination parameters $\boldsymbol{\gamma} \in \mathbb{R}^{27}$ stack $B^2 = 9$ weights per color channel.
Each $\boldsymbol{\gamma}_b \in \mathbb{R}^3$ controls the illumination w.r.t. the red, green and blue channel.

\subsection{Multi-frame Consistent Face Model Learning} \label{sec:learning}

We propose a novel network for consistent multi-frame face model learning.
It consists of $M$ Siamese towers that simultaneously process $M$ frames of the multi-frame image in different streams, see  Fig.~\ref{fig:pipeline}.
Each tower consists of an encoder  that estimates frame-specific parameters and identity feature maps.
Note that the jointly learned geometric identity $\matTheta_s$ and appearance model $(\matTheta_a, \bar{\vecr})$, which are common to all faces, are shared across streams.

\paragraph{Regressed Parameters:}
We train our network in a self-supervised manner based on 
the multi-frame images $\{\setF_{\ell} \}_{\ell=1}^{N}$. %
For each frame $F_{\ell}^{[f]}, \forall f=1:M$ of the multi-frame image $\setF_{\ell}$, we stack the frame-specific parameters regressed by a Siamese tower (see \emph{Parameter Estimation} in Fig.~\ref{fig:pipeline}) in a vector 
$\vecp^{[f]} {=} ( \matR^{[f]}, \vect^{[f]}, \vecgamma^{[f]}, \vecdelta^{[f]})$
that parametrizes rigid pose, illumination and expression.
The frame-independent person-specific identity parameters $\hat{\vecp} {=} (\vecalpha , \vecbeta)$ for the multi-frame image $\setF_{\ell}$  are pooled from all the towers.
We use $\vecp {=} (\hat{\vecp}, \vecp^{[1]},\ldots,\vecp^{[M]})$ to denote all regressed frame-independent and frame-specific parameters of $\setF_{\ell}$.

\paragraph{Per-frame Parameter Estimation Network:}
We employ a convolutional network to extract low-level features.
We then apply a series of convolutions, ReLU, and fully connected layers to regress the per-frame parameters $\vecp^{[f]}$.
We refer to the supplemental document for further details.

\paragraph{Multi-frame Identity Estimation Network:}
As explained in Section~\ref{sec:data}, 
each frame of our multi-frame input exhibits the same face identity under different head poses and expression.
We exploit this information and use a single identity estimation network (see Fig.~\ref{fig:pipeline}) to impose the estimation of common 
identity parameters $\hat{\vecp}$ (shape $\vecalpha$, appearance $\vecbeta$) for all $M$ frames. %
This way, we model a hard constraint on $\hat{\vecp}$ by design.
More precisely, given the frame-specific low-level features obtained by the Siamese networks we apply two additional convolution layers to extract medium-level features.
The resulting $M$ medium-level feature maps are fused into a single multi-frame feature map via average pooling.
Note that the average pooling operation allows us to handle a variable number of inputs. As such, we can perform monocular or multi-view reconstruction at test time, as demonstrated in Sec.~\ref{sec:results}.
This pooled feature map is then fed to an identity parameter estimation network that is based on convolution layers, ReLU, and fully connected layers.
For details, we refer to the supplemental.

\subsection{Loss Functions}
Let $\vecx = (\vecp, \matTheta)$ denote the regressed parameters $\vecp$ as well as the learnable network weights $\matTheta=(\matTheta_s,\matTheta_a, \bar{\vecr})$. Note, $\vecx$ is fully learned during training, whereas the network infers only $\vecp$ at test time.
Here, $\vecp$ is parameterized by the trainable weights of the network.
To measure the reconstruction quality during mini-batch gradient descent, we employ the following loss function: %
\begin{align}
\mathcal{L}(\vecx)
&=  \label{eq:lossdata} 
\lambda_\text{pho} {\cdot} \mathcal{L}_{\text{pho}}(\vecx) + \lambda_\text{lan} {\cdot} \mathcal{L}_{\text{lan}}(\vecx) +\\
&
\lambda_\text{smo} {\cdot}\mathcal{L}_{\text{smo}}(\vecx) + \lambda_\text{spa} {\cdot} \mathcal{L}_{\text{spa}}(\vecx) + \lambda_\text{ble} {\cdot}\mathcal{L}_{\text{ble}}(\vecx)
\enspace{,} \label{eq:lossreg}
\end{align}
which is based on two data terms \eqref{eq:lossdata} and three regularization terms \eqref{eq:lossreg}. %
We found the weights $\lambda_\bullet$ empirically and kept them fixed in all experiments, see supplemental document for details.

\paragraph{Multi-frame Photometric Consistency:}
One of the key contributions of our approach is to enforce multi-frame consistency of the shared identity parameters $\mathbf{\hat p}$.
This can be thought of as solving model-based non-rigid structure-from-motion (NSfM) on each of the multi-frame inputs during training.
We do this by imposing the following photometric consistency loss with respect to the frame $F^{[f]}$:
\begin{align*}
\mathcal{L}_{\text{pho}}( \vecx )
= 
 \sum_{f =1}^M
{
	\sum_{i=1}^{|\mathcal{\hat V}|}
	{ 
		\big|\big|  F^{[f]}( \mathbf{u}_i( \mathbf{p}^{[f]}, \mathbf{\hat p} ) ) - \mathbf{c}_i( \mathbf{p}^{[f]}, \mathbf{\hat p} ) \big|\big|_2^2
	}
}
\enspace{.}
\end{align*}
Here, with abuse of notation, we use $\mathbf{u}_i$ to denote the projection of the $i$-th vertex into screen space, $\mathbf{c}_i$ is its rendered color, and $\mathcal{\hat V}$ is the set of all visible vertices, as determined by back-face culling in the forward pass.
Note that the identity related parameters $\mathbf{\hat p}$ 
are shared across all frames in $\setF$.
This enables a better disentanglement of illumination and appearance, since 
only the illumination and head pose are allowed to change across the frames.

\paragraph{Multi-frame Landmark Consistency:}
To better constrain the problem, we also employ a sparse 2D landmark alignment constraint.
This is based on a set %
of $66$ automatically detected 2D feature points $\mathbf{s}_i^{[f]} \in \mathbb{R}^2$ \cite{SaragLC2011,SaragihLC11a} in each frame $F^{[f]}$.
Each feature point $\mathbf{s}_i^{[f]}$ comes with a confidence $\text{c}_{i}^{[f]}$, so that we use the loss
\begin{align*}
\mathcal{L}_{\text{lan}}( \vecx ) 
=
\sum_{f =1}^M
{
	\sum_{ i=1}^{66}
	{
		\text{c}^{[f]}_{i} \cdot \big|\big| \mathbf{s}_i^{[f]} - \mathbf{u}_{\mathbf{s}_i}(\mathbf{p}^{[f]}, \mathbf{\hat p}) \big|\big|_2^2
	}
	\enspace{.}
}
\end{align*}
Here, $\mathbf{u}_{\mathbf{s}_i} \in \mathbb{R}^{2}$ is the 2D position of the $i$-th mesh feature point in screen space.
We use sliding correspondences, akin to \cite{tewari2018self}.
Note, the position of the mesh landmarks depends both on the predicted per-frame parameters $\mathbf{p}^{[f]}$ and the shared identity parameters $\mathbf{\hat p}$.

\paragraph{Geometry Smoothness on Graph-level:}
We employ a linearized membrane energy \cite{Botsch:2008} to define a first-order geometric smoothness prior on the displacements $\mathbf{t}_i(\mathbf{ \hat p }) = \vecg_i(\mathbf{ \hat p }) - \bar{\vecg}_i $  of the deformation graph nodes  
\begin{equation}
	\mathcal{L}_{\text{smo}}(\vecx)
	=
	\sum_{i =1}^{|\setG|}
	{
		\sum_{j \in \setN_i}{ \big|\big| \mathbf{t}_i(\hat{\vecp}) - \vect_j(\hat{\vecp}) \big|\big|_2^2}
	}
	\enspace{,}
\end{equation}
where $\mathcal{N}_i$ is the set of nodes that have a skinned vertex in common with the $i$-th node.
Note, the graph parameterizes the geometric identity, i.e., it only depends on the shared identity parameters $\mathbf{ \hat p }$.
This term enforces smooth deformations of the parametric shape and leads to higher quality reconstruction results.

\paragraph{Appearance Sparsity:}
In our learned face model, skin appearance is parameterized on a per-vertex basis.
To further constrain the underlying intrinsic decomposition problem, 
we employ a local per-vertex spatial reflectance sparsity prior as in \cite{meka2016,BSTSPP14}, defined as follows
\begin{equation}
	\mathcal{L}_{\text{spa}}(\vecx)
	 =
	 \sum_{i =1}^{|\setV|}
	 {
	 	\sum_{j \in \mathcal{N}_i}
	 	{
	 		w_{ij} \cdot \big|\big| \mathbf{r}_i(\mathbf{\hat p}) - \mathbf{r}_j(\mathbf{\hat p}) \big|\big|_2^p
	 	}
	 }
	 \enspace{.}
\end{equation}
The per-edge weights $w_{ij}$ model the similarity of neighboring vertices in terms of chroma and are defined as
$$w_{ij}=\exp{\big[ -\eta \cdot ||\mathbf{h}_i(\mathbf{\hat p}_{\text{old}}) - \mathbf{h}_j(\mathbf{\hat p}_{\text{old}}) ||_2 \big]} \enspace{.}$$
Here, $\mathbf{h}_i$ is the chroma of $\mathbf{c}_i$ and $\mathbf{\hat p}_{\text{old}}$ denotes the parameters predicted in the last forward pass.
We fix $\eta = 80$ and $p = 0.9$ for training.

\paragraph{Expression Regularization:}
To prevent over-fitting and enable a better learning of the identity basis, we regularize the magnitude of the expression parameters $\vecdelta$:
\begin{equation}
\mathcal{L}_{\text{ble}}( \vecx )
=
\sum_{f =1}^M
{
	\sum_{u = 1}^{|\boldsymbol{\vecdelta}^{[f]}|}
	{
		\Big( \frac{\vecdelta_{u}^{[f]}}{ \sigma_{\vecdelta u}} \Big)^2
	}
}
\enspace{.}
\end{equation}
Here, $\vecdelta_{u}^{[f]}$ is the $u$-th expression parameter of frame $f$, and $\sigma_{\vecdelta u}$ is the corresponding standard deviation computed based on Principal Component Analysis (PCA).

\section{Results} \label{sec:results}

\begin{figure}
	\centering
	\includegraphics[width=\linewidth]{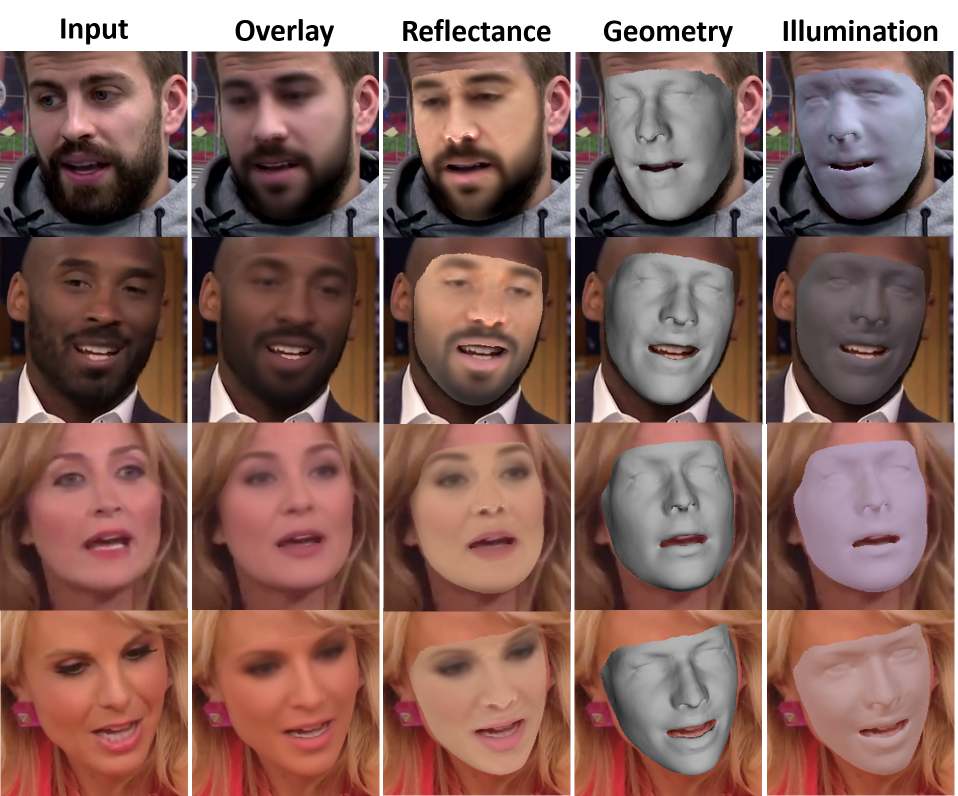}
	\caption{Our approach produces high-quality monocular reconstructions of facial geometry, reflectance and illumination by learning an optimal model from in-the-wild data. This enables us to also reconstruct facial hair and makeup.%
		}
		\vspace{-0.2cm}
	\label{fig:FMLSubjective}
\end{figure}

\begin{figure}
	\centering
	\includegraphics[width=\linewidth]{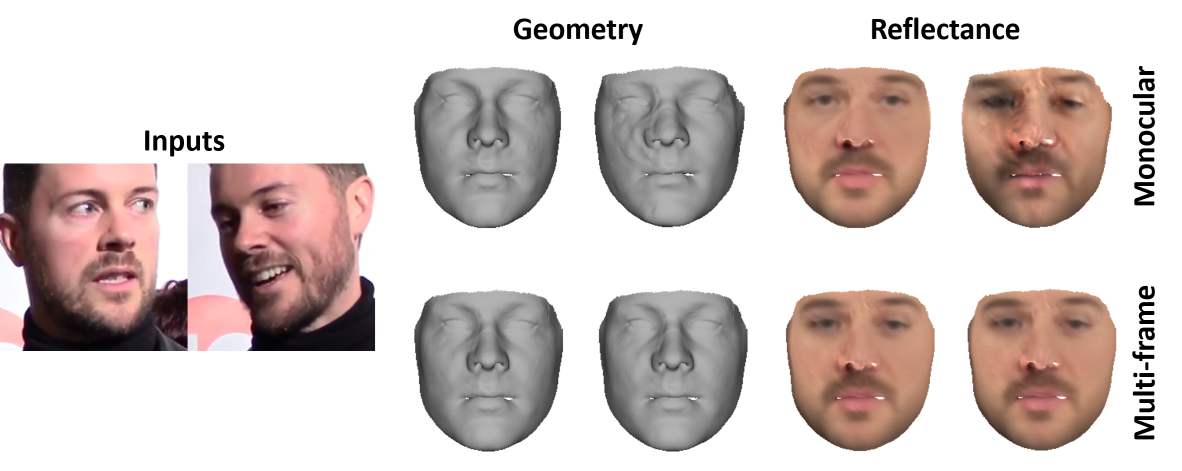}
	\caption{Monocular vs.~multi-frame reconstruction. For clarity, all results are shown with a frontal pose and neutral expression. Multi-view reconstruction improves consistency and quality especially in regions which are occluded in one of the images.}
 \vspace{-0.5cm}
	\label{fig:Mono_vs_Multi}
\end{figure}
We show qualitative results reconstructing geometry, reflectance and scene illumination from monocular images in Fig.~\ref{fig:FMLSubjective}.
As our model is trained on a large corpus of multi-view images, it generalizes well to different ethnicities, even in the presence of facial hair and makeup.
We implement and train our networks in TensorFlow \cite{tensorflow2015-whitepaper}.
We pre-train the expression model and then train the full network end-to-end.
After convergence, the network is fine-tuned using a larger learning rate for reflectance.
We empirically found that this training strategy improves the capture of facial hair, makeup and eyelids, and thus the model's generalization.
Our method can also be applied to multi-frame reconstruction at test time.
Fig.~\ref{fig:Mono_vs_Multi} shows that feeding two images simultaneously improves the consistency and quality of the obtained 3D reconstructions when compared to the monocular case.
Please note that we can successfully separate identity and reflectance due to our novel Orthogonal Complement Layer (OCL).
For the experiments shown in the following sections, we trained our network on $M=4$ multi-frame images and used only one input image at test time, unless stated otherwise.
Our networks take around 30\,hours to train.
Inference takes only 5.2\,ms on a Titan Xp.
More details, results, and experiments can also be found in the supplemental document and video\footnote{\url{http://gvv.mpi-inf.mpg.de/projects/FML19}}.

\begin{figure}
	\centering
	\includegraphics[width=\linewidth]{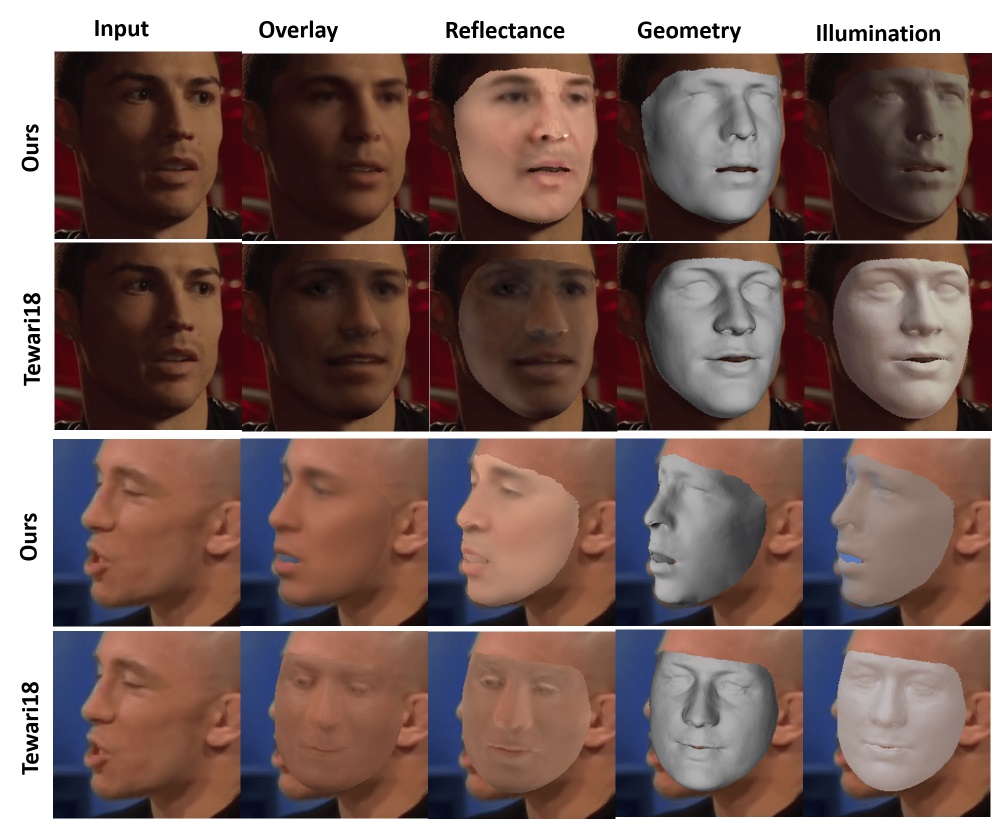}
	\caption{Comparison to Tewari et al.~\cite{tewari2018self}. Multi-frame based training improves illumination estimation. Our approach also outperforms that of Tewari et al. under large poses.}
	\label{fig:MOFAPP}
\vspace{-0.6cm}
\end{figure}

\begin{figure}
	\centering
	\includegraphics[width=\linewidth]{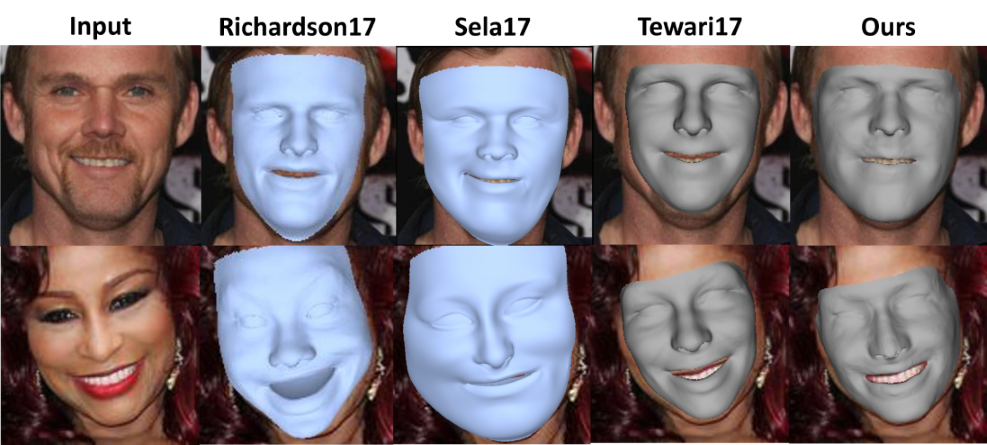}
	\caption{Comparison to \cite{Richardson_2017_CVPR,sela2017unrestricted,tewari17MoFA}. These approaches are constrained by the (synthetic) training corpus and/or underlying 3D face model. Our optimal learned model produces more accurate results, since it is learned from a large corpus of real images.}
	\label{fig:SelarRichardson}
	\vspace{-0.3cm}
\end{figure}

\begin{figure}
	\centering
	\includegraphics[width=0.9\linewidth]{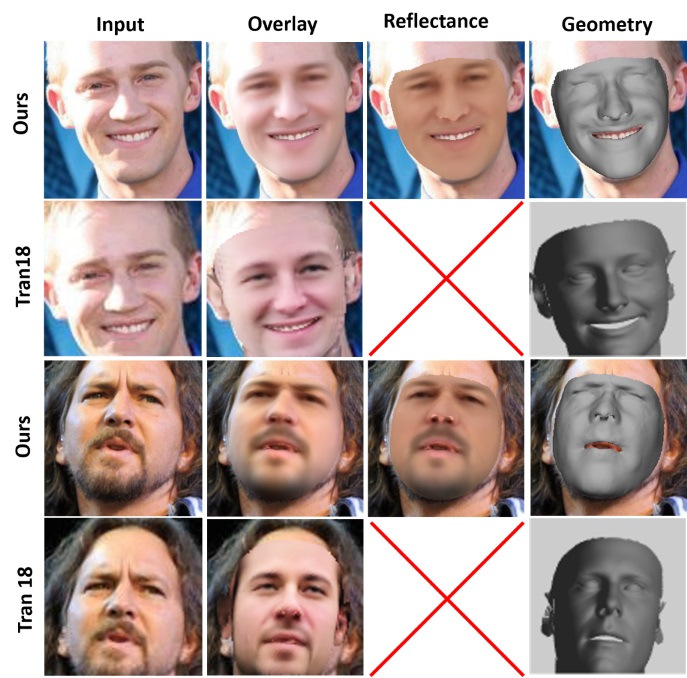}
	\caption{In contrast to Tran et al.~\cite{Tran2018b}, we estimate better geometry and separate reflectance from illumination.
		Note, the approach of Tran et al.~does not disentangle reflectance and shading.}  
	\label{fig:Tran}
	\vspace{-0.3cm}
\end{figure}

\begin{table*}
 \caption{\label{tab:quant_ours}
  Geometric reconstruction error on the BU-3DFE dataset \cite{Yin06}. Our approach produces higher quality results than the current state of the art. The approach of Tewari et al.~\cite{tewari17MoFA} does not generalize to the $\pm45$ degree head poses contained in this dataset.}
\vspace{0.25em}
\footnotesize
\centering
\renewcommand*{\arraystretch}{1.2}
\begin{tabular}{ | c | c | c | c | c | c | c | c | c | c |}
		   \hline
      	   & \multicolumn{5}{|c|}{Ours} & \cite{tewari2018self} Fine & \cite{tewari2018self} Coarse & \cite{tewari17MoFA} \\ \hline
            Train & M = 1 & M = 2 & M = 4 & M = 2 & M = 4 &  & & 
           \\ \hline 
             Test & M = 1 & M = 1 & M = 1 & M = 2 & M = 2 & & & 
           \\ \hline 
    Mean  & 1.92\,mm & 1.82\,mm & 1.76\,mm & 1.80\,mm & \textbf{1.74\,mm} & 1.83\,mm &  1.81\,mm &  3.22\,mm \\ 
    SD    & 0.48\,mm & 0.45\,mm & 0.44\,mm & 0.46\,mm & \textbf{0.43\,mm} & 0.39\,mm &  0.47\,mm &  0.77\,mm \\ \hline
  \end{tabular}
  \label{tab:comparison_BU}
  \vspace{-0em}
\end{table*}
\begin{table*}
  \caption{\label{tab:comparison_fw}
  Geometric error on FaceWarehouse \cite{Cao:2014}.
  Our approach competes with \cite{tewari2018self} and \cite{tewari2018pami}, and outperforms \cite{tewari17MoFA} and \cite{KimZTTRT17}.
  Note, in contrast to these approaches, ours does not require a precomputed face model during training, but learns it from scratch.
  It comes close to the off-line high-quality approach of~\cite{GarriZCVVPT2016}, while being orders of magnitude faster and not requiring feature detection.
  }
  \vspace{0.25em}
\footnotesize
\centering
\renewcommand*{\arraystretch}{1.2}
\begin{tabular}{ | c | c | c | c | c | c | c | c | c | }
		   \hline
    		  & \multicolumn{1}{|c|}{\textbf{Ours}} & \multicolumn{5}{|c|}{Others} & \\ \hline
    		   & \multicolumn{1}{|c|}{Learning} & \multicolumn{4}{|c|}{Learning} & \multicolumn{1}{|c|}{Optimization} & \multicolumn{1}{|c|}{Hybrid} \\ \hline
      	   &  & \cite{tewari2018self} Fine & \cite{tewari2018self} Coarse & \cite{tewari17MoFA} & \cite{KimZTTRT17} & \cite{GarriZCVVPT2016} & \cite{tewari2018pami}\\ \hline
    Mean  & \textbf{1.90\,mm} & 1.84\,mm &  2.03\,mm &  2.19\,mm &  2.11\,mm & \textbf{1.59\,mm} & 1.87\,mm\\ 
    SD     & \textbf{0.40\,mm} & 0.38\,mm &  0.52\,mm &  0.54\,mm &  0.46\,mm & \textbf{0.30\,mm} & 0.42\,mm\\ \hline
    Time  & \textbf{5.2\,ms} & \textbf{4\,ms} &  \textbf{4\,ms} &  \textbf{4\,ms} & \textbf{4\,ms} & 120\,s & 110\,ms \\ \hline
  \end{tabular}
  \label{tab:quant_fw}
  \vspace{-1em}
\end{table*}
\begin{figure}
	\centering
	\includegraphics[width=0.9\linewidth]{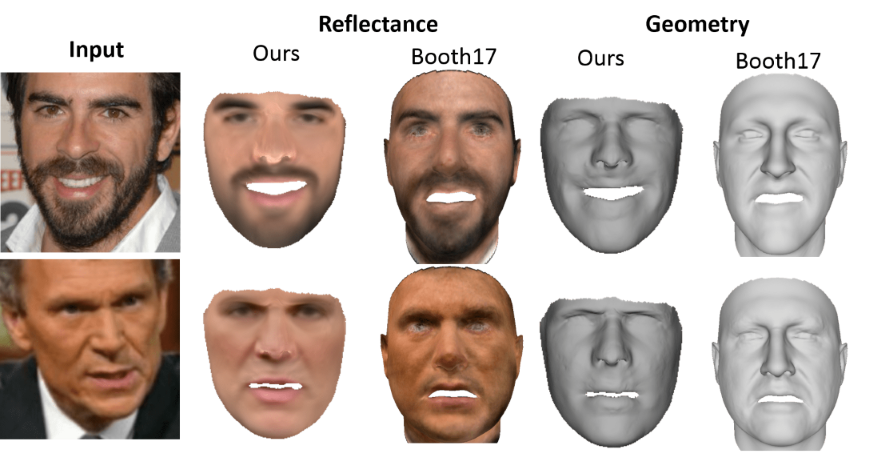}
	\caption{In contrast to the texture model of Booth et al. \cite{Booth_2017_CVPR} that contains shading, our approach estimates a reflectance model.}
	\label{fig:Booth}
	\vspace{-0.6cm}
\end{figure}

\subsection{Comparisons to Monocular Approaches}
State-of-the-art monocular reconstruction approaches that rely on an existing face model \cite{tewari17MoFA} or synthetically generated data \cite{sela2017unrestricted,Richardson_2017_CVPR} during training, do not generalize well to faces outside the span of the model. As such, they can not handle facial hair, makeup, and unmodeled expressions, see Fig.~\ref{fig:SelarRichardson}.
Since we train our models on in-the-wild videos, we can capture these variations and thus generalize better in such challenging cases.
We also compare to the refinement based approaches of \cite{tewari2018self,Tran2018b}. %
Tran et al.~\cite{Tran2018b} (see Fig~\ref{fig:Tran}) refine a 3DMM \cite{Blanz1999} based on in-the-wild data. 
Our approach produces better geometry without requiring a 3DMM and, contrary to \cite{Tran2018b}, it also separates albedo from illumination.
The approach of Tewari et al.~\cite{tewari2018self} (see Fig~\ref{fig:MOFAPP}) requires a 3DMM \cite{Blanz1999} as input and only learns shape and reflectance correctives.
Since they learn from monocular data, their correctives are prone to artifacts, especially when occlusions or extreme head poses exist. 
In contrast, our approach learns a complete model from scratch based on multi-view supervision, thus improving robustness and reconstruction quality.
We also compare to \cite{Booth_2017_CVPR}, which only learns a texture model, see Fig.~\ref{fig:Booth}. 
In contrast, our approach learns a model that separates albedo from illumination. 
Besides, their method needs a 3DMM~\cite{Blanz1999} as initialization, while we start from a single constantly colored mesh and learn all variation modes (geometry and reflectance) from scratch.

\subsection{Quantitative Results}
We also evaluate our reconstructions quantitatively on a subset of the BU-3DFE dataset \cite{Yin06}, see Tab.~\ref{tab:comparison_BU}. 
This dataset contains images and corresponding ground truth geometry of multiple people performing a variety of expressions.
It includes two different viewpoints.
We evaluate the importance of multi-frame training in the case of monocular reconstruction using per-vertex root mean squared error based on a pre-computed dense correspondence map.
The lowest error is achieved with multi-view supervision during training, in comparison to monocular input data.
Multi-view supervision can better resolve depth ambiguity and thus learn a more accurate model.
In addition, the multi-view supervision also leads to a better disentanglement of reflectance and shading.
We also evaluate the advantage of multi-frame input at test time. 
When both images corresponding to a shape are given, we consistently obtain better results. 
Further, our estimates are better than the state-of-the-art approach of \cite{tewari2018self}. 
Since \cite{tewari2018self} refine an existing 3DMM only using monocular images during training, it cannot resolve depth ambiguity well. Thus, it does not improve the performance compared to their coarse model on the $\pm 45$ degree poses of BU-3DFE \cite{Yin06}.
Similar to previous work, we also evaluate monocular reconstruction on 180 meshes of FaceWarehouse \cite{Cao:2014}, see Tab.~\ref{tab:comparison_fw}.
We perform similar to the 3DMM-based state-of-the-art.
Note that we do not use a precomputed 3DMM, but learn a model from scratch during training, unlike all other approaches in this comparison.
For this test, we employ a model learned starting from an asian mean face, as FaceWarehouse mainly contains asians.
Our approach is agnostic to the mean face chosen and thus allows us this freedom.

\section{Conclusion \& Discussion}
We have proposed a self-supervised approach for joint multi-frame learning of a face model and a 3D face reconstruction network.
Our model is learned from scratch based on a large corpus of in-the-wild video clips without available ground truth.
Although we have demonstrated compelling results by learning from in-the-wild data, such data is often of low resolution, noisy, or blurred, which imposes a bound on the achievable quality.
Nevertheless, our approach already matches or outperforms the state-of-the-art in learning-based face reconstruction.
We hope that it will inspire follow-up work and that multi-view supervision for learning 3D face reconstruction will receive more attention.

\vspace{-0.5cm}
\let\thefootnote\relax\footnotetext{
	\textbf{Acknowledgements:}
We thank True-VisionSolutions Pty Ltd for providing the 2D face tracker, and the authors of \cite{Booth_2017_CVPR, Richardson_2017_CVPR, sela2017unrestricted, Tran2018b} for the comparisons.
We also thank Franziska M{\"u}ller for the video voiceover.
This work was supported by the ERC Consolidator Grant 4DReply (770784), the Max Planck Center for Visual Computing and Communications (MPC-VCC), and by Technicolor.
}
\newpage
{\small
\bibliographystyle{ieee}
\bibliography{fml}
}
\clearpage


\section*{Supplementary material}

# FML: Face Model Learning from Videos
## — Supplemental Document —

Ayush Tewari[1]   Florian Bernard[1]   Pablo Garrido[2]   Gaurav Bharaj[2]   Mohamed Elgharib[1]
Hans-Peter Seidel[1]   Patrick Pérez[3]   Michael Zollhöfer[4]   Christian Theobalt[1]

[1]MPI Informatics, Saarland Informatics Campus   [2]Technicolor   [3]Valeo.ai   [4]Stanford University

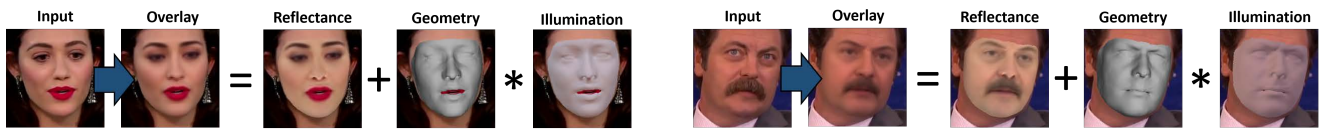

Figure 1. We propose multi-frame self-supervised training of a deep network based on in-the-wild video data for jointly learning a face model and 3D face reconstruction. Our approach successfully disentangles facial shape, appearance, expression, and scene illumination.

In this supplemental document, we provide more details on the network architecture and the empirically determined weights in the employed loss function. We also show more qualitative and quantitative comparisons and discuss limitations of our approach. In addition, we describe how to extract statistics from the learned model and visualize its modes. A large amount of qualitative results can also be found on the supplemental webpage[1].

## 1. Network Details

We provide further details of the feature extraction, shared identity and parameter estimation network in Tab. 1, 2 and 3, respectively. Overall, our network has 124M parameters. Note that a subset of these parameters includes the learned geometry and appearance model. We train our networks on commodity Titan Volta GPUs.

### 1.1. Visualizing the Modes of Variation

While the learned model is an optimal basis for the monocular face reconstruction task, it does not allow for an intuitive analysis of the most prominent modes of variation observed in the data. However, we can easily reparameterize the learned model and construct a new representation using e.g. Principle Component Analysis (PCA). More specifically, we compute PCA on 3D reconstructions obtained by our approach for over 10k images of our training set. Note, our approach is trained in a self-supervised manner without requiring ground truth in the form of dense geometry and appearance annotations. The new parametrization allows us

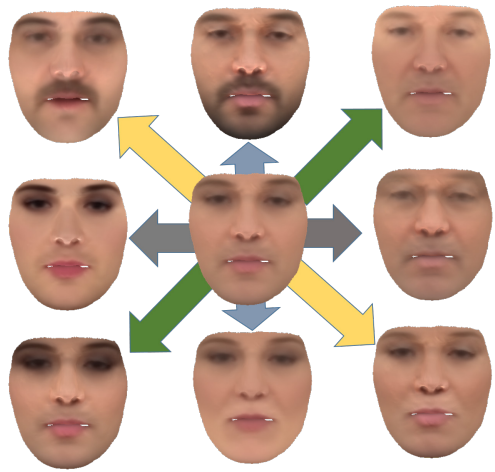

Figure 2. Visualization of the Reflectance Model. We show four of the learned modes.

to build a statistical face model of facial identity and appearance, as in [1], but based on in-the-wild video data, see Fig. 2, 3 and 4. Our model learns global variation modes that roughly correspond to gender (see Fig. 2) as well as local variation modes, such as nose and eye deformations (see Fig. 3). The visualization also shows the separation between the learned shape identity model and the expression model.

## 2. Weights of the Energy

We found the weights in our energy empirically and kept them fixed in all experiments: $\lambda_{\text{pho}} = 1.6/|\bar{\mathcal{V}}|, \lambda_{\text{lan}} =$

---
[1]http://gvv.mpi-inf.mpg.de/projects/FML19

Table 1. Feature Extractor Network Details. ↑ means that the input is taken from the layer in the row above.

| Input | Layers | Activation Shape | Siamese | Output |
|---|---|---|---|---|
| Image (240, 240, 3) | Conv2D (kernel 11x11, stride 4) + ReLU | (60, 60, 96) | Yes | *unnamed* |
| ↑ | MaxPool (kernel 3x3, stride 2) | (29, 29, 96) | n/a | *unnamed* |
| ↑ | Conv2D (kernel 5x5, stride 1) + ReLU | (14, 14, 256) | Yes | *unnamed* |
| ↑ | Conv2D (kernel 3x3, stride 1) + ReLU | (14, 14, 384) | Yes | *lowFeatures$_f$* |
| ↑ | Conv2D (kernel 3x3, stride 2) + ReLU | (7, 7, 256) | Yes | *unnamed* |
| ↑ | Conv2D (kernel 3x3, stride 2) + ReLU | (4, 4, 256) | Yes | *mediumFeatures$_f$* |

Table 2. Shared Identity Network Details. ↑ means that the input is taken from the layer in the row above.

| Input | Layers | Activation Shape | Siamese | Output |
|---|---|---|---|---|
| lowFeatures$_{0\ldots M}$ | Concat | (M, 4, 4, 256) | n/a | *unnamed* |
| ↑ | MeanPool | (4, 4, 256) | n/a | *unnamed* |
| ↑ | Conv2D (kernel 3x3, stride 1) + ReLU | (4, 4, 384) | No | *unnamed* |
| ↑ | Conv2D (kernel 3x3, stride 1) + ReLU | (4, 4, 256) | No | *unnamed* |
| ↑ | Fully Connected + ReLU | (1000, 1) | No | *unnamed* |
| ↑ | Fully Connected + ReLU | (1000, 1) | No | *unnamed* |
| ↑ | Fully Connected | (500 + 500, 1) | No | *shapeParam + reflectanceParam* |

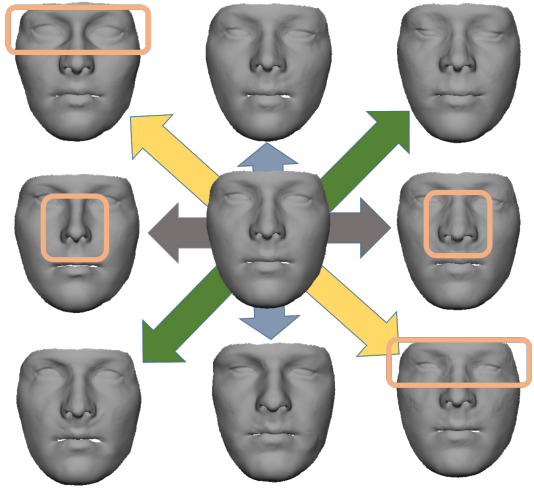

Figure 3. Visualization of the Shape Model. We show four of the learned modes.

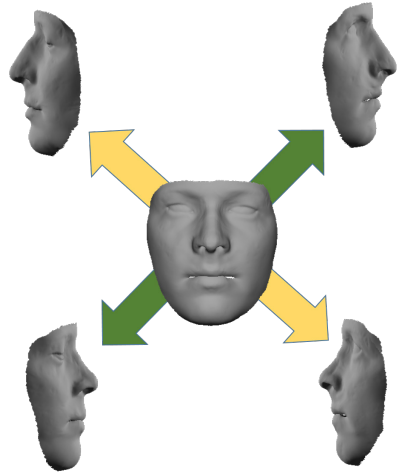

Figure 4. Visualization of the Shape Model. We show two of the learned modes from a side view.

4.7, $\lambda_{\text{smo}} = 0.001$, $\lambda_{\text{spa}} = 1e-7$, $\lambda_{\text{ble}} = 1e-8$.

## 3. Results

In the following, we show more detailed qualitative and quantitative evaluations of our approach.

**Monocular vs multi-frame reconstruction** Figure 5 shows the advantage of using multiple frames at test time. Multi-frame reconstruction improves overall consistency of the estimated 3D face and resolves ambiguity due to occlusions that are present in one of the images. Such ambiguities cannot be resolved completely when feeding only a single image to the network. Still, our network is able to obtain plausible facial identity for the monocular case thanks to our multi-frame based training.

**Comparison to state-of-the-art methods** Fig. 6, 7, 8 and 9 show more comparisons to related state-of-the-art approaches for monocular 3D face reconstruction. Our multi-frame based training succeeds in reconstructing 3D faces for images with large poses and harsh yet low-frequency illumination, as shown in Fig. 6. The method of Tewari et al. [8] is unable to deal with these cases, as it is trained on monocular images. Fig. 7 shows that our approach also generalizes well to facial identities having beards and non-average faces thanks to the learning of the optimal model from in-the-wild data. On the contrary, methods relying on synthetic data [5, 7] and/or an underlying 3DMM [9], fail to generalize to novel identities not explained by the 3D model or training corpus. Our approach not only estimates 3D faces from challenging in-the-wild images, but

Table 3. Parameter Estimation Network Details. ↑ means that the input is taken from the layer in the row above.

| Inputs | Layers | Activation Shape | Siamese | Output |
|---|---|---|---|---|
| shapeParam, reflectanceParam | Fully Connected + ReLU + Reshape | (14, 14, 1) | No | *unnamed* |
| ↑ | Conv2D (kernel 3x3, stride 1) + ReLU | (14, 14, 384) | No | *unnamed* |
| ↑, lowFeatures$_f$ | Concat | (14, 14, 768) | n/a | *unnamed* |
| ↑ | Conv2D (kernel 3x3, stride 1) + ReLU | (14, 14, 384) | Yes | *unnamed* |
| ↑ | Conv2D (kernel 3x3, stride 1) + ReLU | (14, 14, 384) | Yes | *unnamed* |
| ↑ | Conv2D (kernel 3x3, stride 1) + ReLU | (14, 14, 256) | Yes | *unnamed* |
| ↑ | MaxPool(kernel 3x3, stride 2) | (6, 6, 256) | Yes | *unnamed* |
| ↑ | Fully Connected + ReLU | (2048, 1) | Yes | *unnamed* |
| ↑ | Fully Connected | (6 + 64 + 27, 1) | Yes | rigid$_f$ + expressionParam$_f$ + illuminationParam$_f$ |

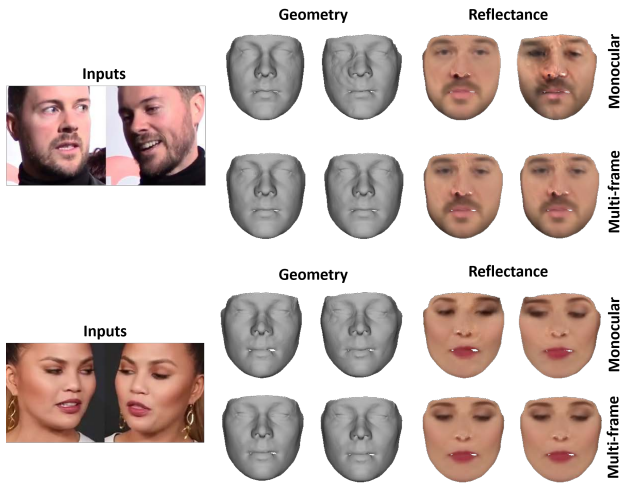

Figure 5. Monocular vs. multi-frame reconstruction. Multi-view reconstruction improves consistency and reconstruction quality, especially for regions occluded in one of the images. Note that all results are shown with a frontal pose and neutral expression for comparison purposes.

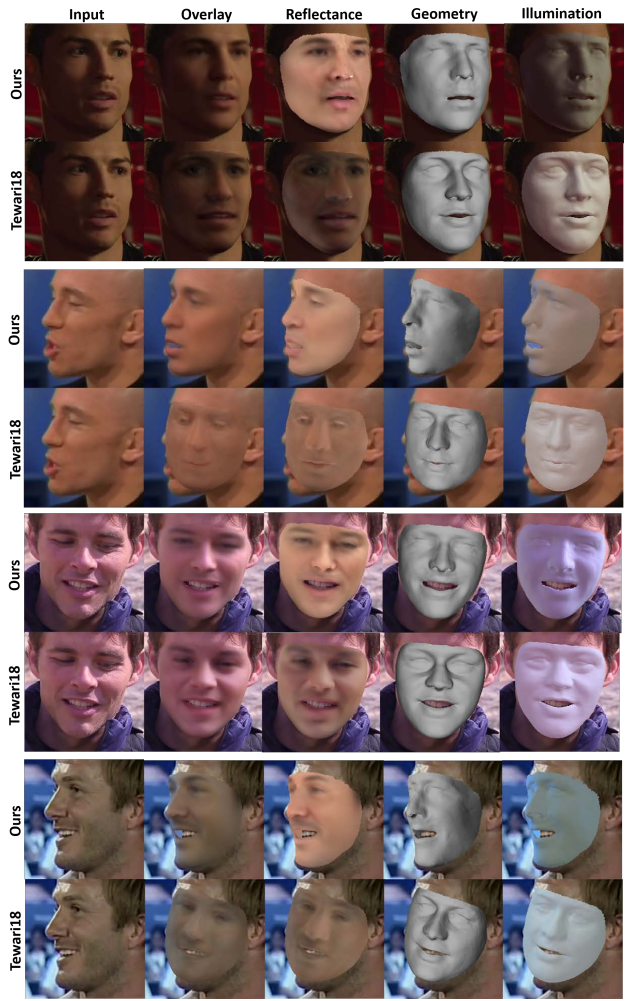

Figure 6. Comparison to Tewari et al. [8]. Multi-frame based training improves illumination estimation. Our approach also outperforms that of Tewari et al. when the face is largely occluded.

also successfully disentangles facial geometry, reflectance and scene illumination. Fig. 8 and 9 show that we obtain a fairly clean reflectance estimation, up to a small global scaling factor. Current state-of-the-art methods, on the contrary, only estimate facial texture that bakes in shading effects [11, 2]. We remark that our approach learns a reflectance model from scratch using only a colored template mesh, whereas the method of Booth et al. [2] require a 3DMM as initialization to learn a texture model, see Fig. 9.

**Quantitative evaluations** We quantitatively evaluate the photometric error of our approach on 1000 images of the CelebA dataset [4], see Fig. 11 and Tab. 4. We achieve lower errors when using larger models for shape and geometry. We also obtain lower errors compared to the 3DMM-based optimization approach presented in [3]. This demonstrates better generalization capabilities of our learned shape and appearance models to in-the-wild images, compared to the fixed face model [1] used by [3].

Fig. 12 shows the quantitative evaluation of our geometry reconstruction on 324 images of the BU-3DFE dataset [12]. Training on multiple frames consistently improves reconstruction quality. Multi-frame reconstruction with two

Table 4. Average photometric error (R,G,B $\in [0, 255]$) over 1000 images of the CelebA[4] dataset. Size refers to the number of vectors in our learned shape and appearance models. Larger models lead to lower errors. Our method outperforms [3] which reconstructs faces using an existing face model [1].

| | Ours | | | | | [3] |
|---|---|---|---|---|---|---|
| Size | 0 | 10 | 50 | 125 | 500 | 80 |
| Mean | 32.54 | 23.13 | 21.27 | 20.71 | **20.65** | 21.95 |
| SD | 8.88 | 6.66 | 6.15 | 6.04 | 6.04 | 5.60 |

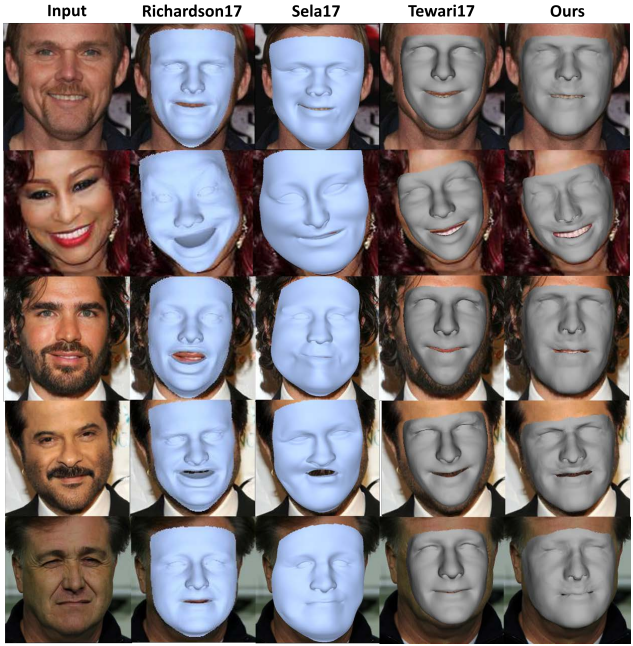

Figure 7. Comparison to [6, 7, 9]. These approaches are constrained by the (synthetic) training corpus and/or underlying 3D face model. Our optimal learned model produces more accurate results, since it is learned from a large corpus of real images.

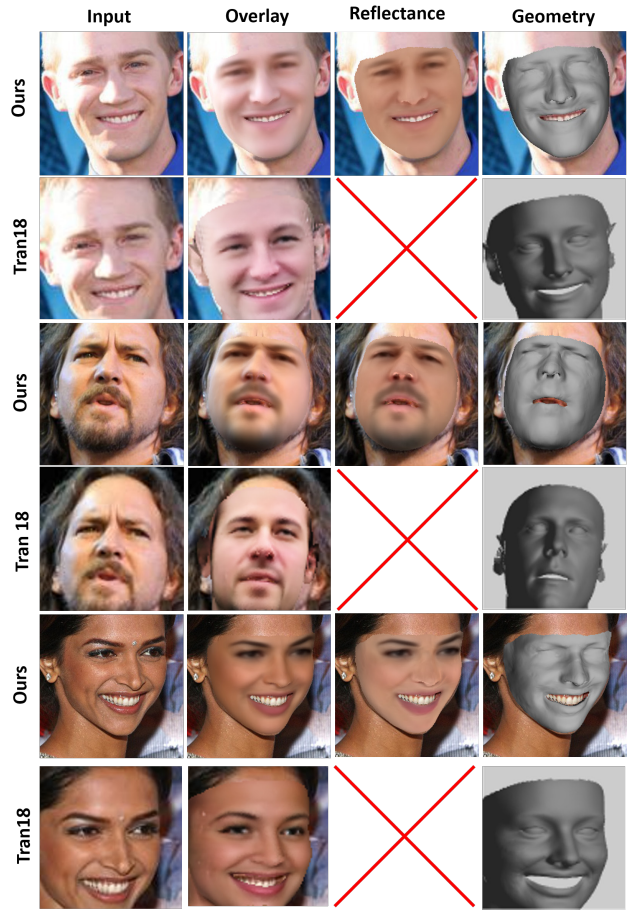

Figure 8. In contrast to Tran et al. [10], we estimate better geometry and separate reflectance from illumination. Note, the approach of Tran et al. does not disentangle reflectance and shading.

images at test time also increases reconstruction quality compared to the monocular reconstruction case. We remark that our approach outperforms that of Tewari et al. [9, 8] on this dataset.

**Evaluation of different pooling techniques** We use average pooling for fusing features extracted from the multi-frame images. We evaluate the performance of multi-frame reconstruction using max pooling as well. With $M = 4$ at training and $M = 2$ at test time, both pooling techniques lead to very similar geometric reconstruction errors on BU-3DFE (difference is less than $0.004$mm). However, other pooling techniques could further improve our results.

## 4. Limitations

In this paper, we have proposed a multi-frame self-supervised deep learning approach that jointly learns a 3D face model (3D geometry and facial identity) and reconstructs 3D faces from in-the-wild videos. Although we have shown compelling results, our approach still has a few limitations that can be addressed in follow-up work, see Fig. 10. Overall our approach can deal with large head poses quite well. Still, reconstructing extreme poses is a hard task in itself that challenges all face reconstruction techniques. Occlusions, e.g., by accessories or thick facial hair might adversely impact the reconstruction quality of our approach. Facial hair, such as beards are modeled in the reflectance channel, and thus are not reconstructed in a physically correct manner. Even though our multi-frame supervision approach can obtain quite clean reflectance estimates that are free of shading, there is still a remaining global scale am-

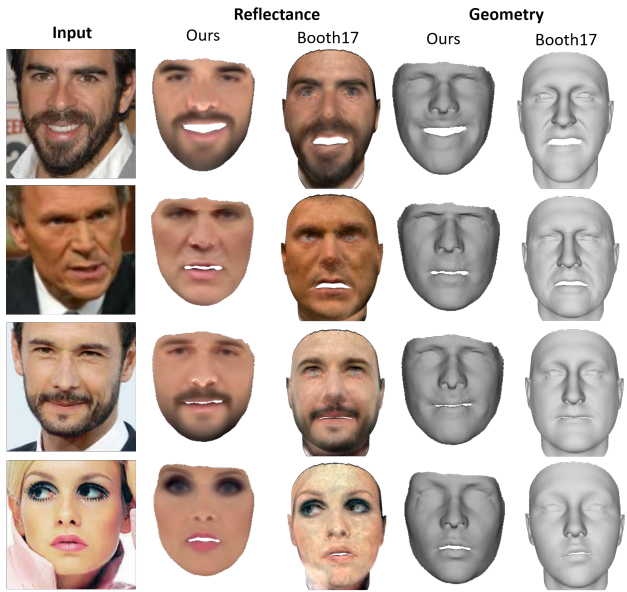

Figure 9. In contrast to the texture model of Booth et al. [2] that contains shading, our approach estimates a reflectance model.

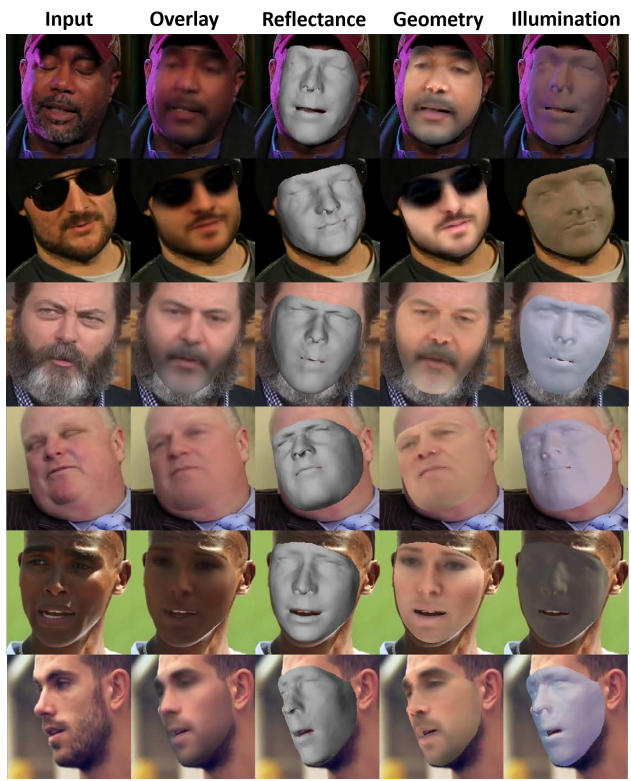

Figure 10. Limitations of our approach. From top to bottom: Extreme illumination conditions, severe occlusions by accessories, thick facial hair, non-average facial shapes, scale ambiguity between illumination and reflectance, and extreme head poses.

biguity between illumination and reflectance. As such, the global skin tone can not be reliably disentangled from the general ambient brightness of the illumination. Strong and colorful directional illumination outside the norm might also harm the estimation of 3D faces. Specular reflections and cast shadows are currently not modeled by our differentiable renderer, and thus they might slightly be baked into the reflectance channel. Non-standard facial shapes challenge our approach. We remark that all of these are difficult settings for almost any face reconstruction technique. Our approach already handles the aforementioned cases quite well by learning from in-the-wild videos without any sort of explicit 3D supervision.

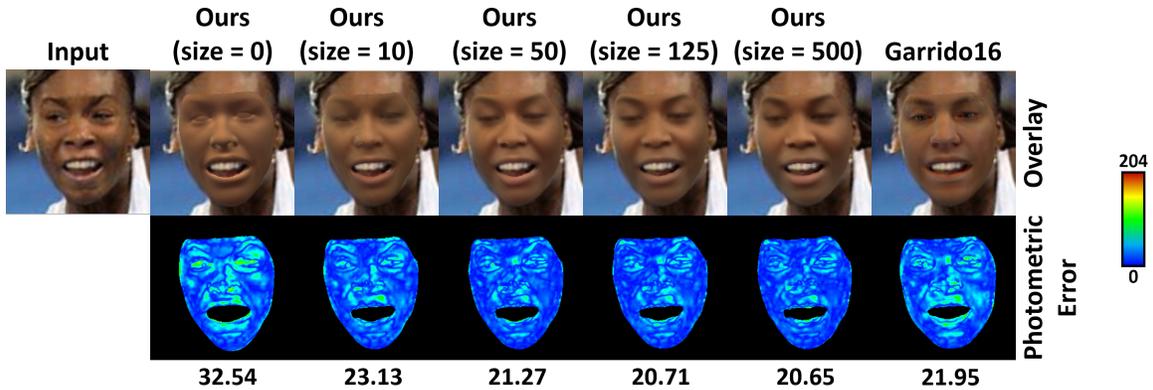

Figure 11. Quantitative evaluation of photometric error on the CelebA [4] dataset. Size is the number of learnable vectors in our shape and appearance models. Our method outperforms [3] which uses an existing model for reconstruction. The numbers are the average photometric errors (R,G,B $\in [0, 255]$) over 1000 images of the CelebA[4] dataset.

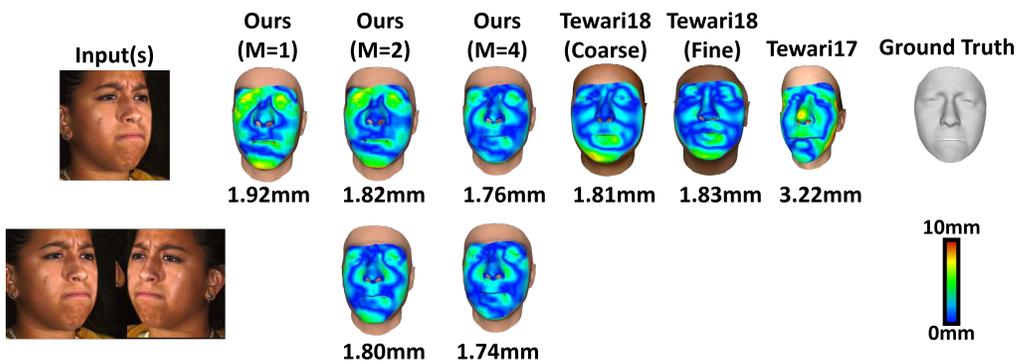

Figure 12. Quantitative evaluation on the BU-3DFE [12] dataset. The numbers are the geometric reconstruction errors averaged over 324 meshes. $M$ is the size of the multi-frame images used at training time. Multi-frame inputs at training and at testing time help in obtaining better reconstructions.


\end{document}